\pdfoutput=1

\documentclass{article}
\usepackage{spconf}
\usepackage{amsmath,amsfonts,amssymb}
\usepackage{graphicx}
\usepackage{xfrac}
\usepackage{tabularx}
\usepackage{footmisc}
\usepackage{siunitx}
\usepackage{booktabs}
\usepackage{float}
\usepackage{hyperref}
\hypersetup{
    colorlinks=true,
    linkcolor=blue, 
    citecolor=black, 
    urlcolor=blue   
}

\usepackage{tikz}
\usetikzlibrary{quotes,arrows.meta}
\usepackage{xcolor}
\definecolor{NavyBlue}{rgb}{0,0,0.5}

\usepackage{scalerel}
\usepackage{tikz}
\usetikzlibrary{svg.path}

\definecolor{orcidlogocol}{HTML}{A6CE39}
\tikzset{
	orcidlogo/.pic={
		\fill[orcidlogocol] svg{M256,128c0,70.7-57.3,128-128,128C57.3,256,0,198.7,0,128C0,57.3,57.3,0,128,0C198.7,0,256,57.3,256,128z};
		\fill[white] svg{M86.3,186.2H70.9V79.1h15.4v48.4V186.2z}
		svg{M108.9,79.1h41.6c39.6,0,57,28.3,57,53.6c0,27.5-21.5,53.6-56.8,53.6h-41.8V79.1z M124.3,172.4h24.5c34.9,0,42.9-26.5,42.9-39.7c0-21.5-13.7-39.7-43.7-39.7h-23.7V172.4z}
		svg{M88.7,56.8c0,5.5-4.5,10.1-10.1,10.1c-5.6,0-10.1-4.6-10.1-10.1c0-5.6,4.5-10.1,10.1-10.1C84.2,46.7,88.7,51.3,88.7,56.8z};
	}
}

\newcommand\orcidicon[1]{\href{https://orcid.org/#1}{\mbox{\scalerel*{
				\begin{tikzpicture}[yscale=-1,transform shape]
				\pic{orcidlogo};
				\end{tikzpicture}
			}{|}}}}

\def\x{{\mathbf x}}

\newif\ifdraft
\drafttrue

\definecolor{orange}{rgb}{1,0.5,0}
\definecolor{pink}{rgb}{0.98, 0.38, 0.5}
\definecolor{darkgreen}{rgb}{0.055, 0.490, 0.016} 

\definecolor{senioryellow}{rgb}{0.86, 0.8, 0.19}
\definecolor{assistantblue}{rgb}{0, 0.133, 0.31}

\ifdraft
 \usepackage[normalem]{ulem}
 \newcommand{\RS}[1]{{\color{red}{\bf RS: #1}}}
 
 \newcommand{\PMN}[1]{{\color{orange}{\bf PMN: #1}}}

\else
 \newcommand{\sout}[1]{}
 \newcommand{\RS}[1]{{\color{red}{}}}
 
 \newcommand{\PMN}[1]{{\color{red}{}}}
 
\fi

\newcommand{\comment}[1]{}

\title{StofNet: Super-resolution Time of Flight Network}
\name{Christopher Hahne$^{\star}$
\qquad 
Michel Hayoz \qquad 
Raphael Sznitman\thanks{This study is funded by the Hasler Foundation grant number 22027.\\
$^{\star}$Corresponding author email: \href{mailto:christopher.hahne@unibe.ch}{\textcolor{blue}{christopher.hahne [ät] unibe.ch}}
}}
\address{ARTORG Center, University of Bern, Switzerland}
\begin{document}
%
\maketitle
\begin{abstract} 
Time of Flight (ToF) is a prevalent depth sensing technology in the fields of robotics, medical imaging, and \mbox{non-destructive} testing. 
Yet, ToF sensing faces challenges from complex ambient conditions making an inverse modelling from the sparse temporal information intractable.
This paper highlights the potential of modern \mbox{super-resolution} techniques to learn varying surroundings for a reliable and accurate ToF detection.
%
Unlike existing models, we tailor an architecture for sub-sample precise semi-global signal localization by combining super-resolution with an efficient residual contraction block to balance between fine signal details and large scale contextual information.
We consolidate research on ToF by conducting a benchmark comparison against six state-of-the-art methods for which we employ two publicly available datasets. This includes the release of our SToF-Chirp dataset captured by an airborne ultrasound transducer. 
Results showcase the superior performance of our proposed StofNet in terms of precision, reliability and model complexity.
Our code is available at \texttt{{\color{NavyBlue}\textmd{\url{{https://github.com/hahnec/stofnet}}}}}.
\end{abstract}
\section{Introduction}
\label{sec:intro}
%
Time of Flight (ToF) is a widely used non-contact depth sensing technology in various fields, including geometric scene reconstruction~\cite{li2019multi,Christensen:2020,Parida:2021,Irie:22,berg22_interspeech}
, robotics~\cite{Tracy:2021,spiel:icra23}, autonomous vehicles~\cite{wan2020deep}, non-destructive testing (e.g., structural health monitoring)~\cite{kurz2005strategies,pearson2017improved,Zonzini:2022} and medical imaging~\cite{hahne2023gulm}. 
For example, the precise detection of wavefront arrival times enables geometric localization of targets, which is also known as multilateration~\cite{astrom2021,berg22_interspeech,spiel:icra23,hahne2023gulm}. 
%
The core concept of ToF relies on measuring the time it takes for a signal carrier, such as light or sound, to travel from a source to a target and back to the receiver, which is also known as Time of Arrival (ToA). 

Despite its wide adoption, ToF sensing still faces challenges such as mitigating the impact of ambient influences (e.g., target shape, temperature, humidity), improving the accuracy and resolution of distance measurements and handling multipath interference, which occurs when the signal reflects off multiple surfaces or objects before reaching the sensor. %
Such ambient conditions can be attributed to the physical emission and reflection properties as well as the surrounding environment (target position, material stiffness, signal penetration, temperature, etc.). Modeling these influences as inverse problems rapidly turns out to be cumbersome. 
%
Instead, the rise of deep neural networks offers new opportunities to address these demands and further enhance the capabilities and robustness of ToF sensing systems. Can recent advances in \mbox{super-resolution} imaging be transferred and equally applied to localization in the temporal domain? What kind of modifications are necessary to ensure enhanced operation considering established methods?

ToF determination breaks down to a localization problem across the temporal domain of a captured one-dimensional \mbox{(1-D)} signal and has been thoroughly studied in the past. Traditional ToA detection includes thresholding~\cite{kurz2005strategies,pearson2017improved,Hahne:22} and cross-correlation methods~\cite{adrian2015acoustic,berg22_interspeech}. The vastly growing field of deep learning offers many considerable architectures for that task~\cite{ronneberger2015u,kuleshov2017audio,ravanelli2018speaker,Zonzini:2022}. Often, these models contain a contraction and expansion resembling the shape of the U letter~\cite{ronneberger2015u}. This trend can also be seen in the acoustic signal processing domain~\cite{ravanelli2018speaker,Zonzini:2022} and particularly in the audio super-resolution field~\cite{kuleshov2017audio}. The general motivation for such U-Net is to establish global context for regions larger than the convolution kernel size. 
While this comes in handy for semantic segmentation, spatial or temporal downscaling carries the risk of losing the capability to recover fine signal details. 
Instead of a bottleneck contraction, advances in efficient image super-resolution networks omit spatial downscaling while expanding the feature channels to learn upsampling from a trailing feature channel shuffle operation~\cite{shi2016real,lim2017enhanced}. On the contrary, these super-resolution networks lack in building large-scale contextual information, which becomes essential among extended signal regions, e.g. to avoid multiple detections for a single target. \par

In this paper, we customize a network based on~\cite{shi2016real,lim2017enhanced} for temporal \mbox{1-D} localization while balancing the above requirements to improve distance accuracy and the reliability of ToF measurements. We compare our architecture against traditional ToF detection techniques~\cite{kurz2005strategies,Hahne:22}, audio super-resolution networks~\cite{kuleshov2017audio,ravanelli2018speaker,Zonzini:2022} and \mbox{1-D} equivalents of models used in computer vision~\cite{shi2016real,lim2017enhanced}.

\begin{figure*}[t]
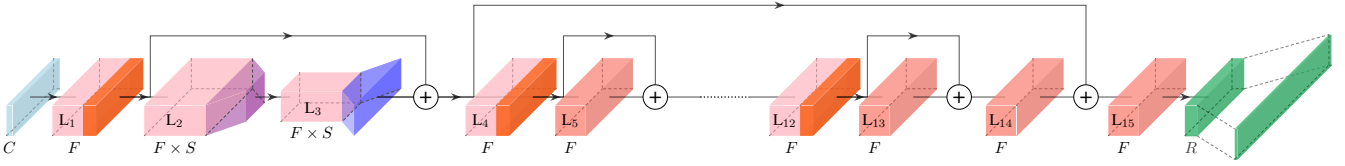

    \begin{minipage}[b]{\linewidth}
        \centerline{\resizebox{\linewidth}{!}{\begin{tikzpicture}

\definecolor{salmon}{RGB}{250, 128, 114} 
\definecolor{lightsalmon}{RGB}{255, 160, 122}
\definecolor{lightpink}{RGB}{255, 182, 193}
\definecolor{PALEGOLDENROD}{RGB}{238, 232, 170}
\definecolor{nephritis}{RGB}{39, 174, 96}
\definecolor{blueviolet}{RGB}{148,0,211}
\definecolor{lightblue}{RGB}{173, 216, 230}
\definecolor{lime}{RGB}{0, 255, 0}

\newcommand{\myopacity}{.75}

\tikzstyle{connection}=[thin,every node/.style={sloped,allow upside down, midway},draw=black,opacity=0.7]

\input{plot_stofnet/layer/Arrow.sty}
\input{plot_stofnet/layer/Ball.sty}
\input{plot_stofnet/layer/Box.sty}
\input{plot_stofnet/layer/TemporalBox.sty}

\pic[shift={(0,0,0)}] at (-15,0,0)
    {BaseBox={
        name=input,
        opacity=\myopacity,
        xlabel=$C$, 
        fill=lightblue, 
        scriptscale=1.5, 
        width=.2,
        height=1,
        depth=4,
        border=none
        }
    };

\draw [connection] (-15,0,0) -- (-13.5,0,0) node{\midarrow};

\pic[shift={(0,0,0)}] at (-13.5,0,0) 
    {ConvReLU={
        name=l1relu,
        opacity=\myopacity,
        caption=L$_1$,
        xlabel=$F$, 
        fill=lightpink,
        scriptscale=1.5, 
        width=1.4285714285714286,
        height=1,
        depth=4,
        border=none
        }
    };

\draw [connection] (l1relu-right) -- (-10.5,0,0) node{\midarrow};

\pic[shift={(0,0,0)}] at (-10.5,0,0) 
    {ContractTemporalReLU={
        name=contract,
        opacity=\myopacity,
        caption=L$_2$,
        xlabel=$F\times S$, 
        fill=lightpink,
        scriptscale=1.5, 
        width=2.857142857142857,
        height=1,
        depth=4,
        border=none
        }
    };

\draw [connection] (contract-right) -- (-6.5,0,0) node{\midarrow};

\pic[shift={(0,0,0)}] at (-6.5,0,0) 
    {ExpandTemporalReLU={
        name=expand,
        opacity=\myopacity,
        caption=L$_3$,
        xlabel=$F\times S$, 
        fill=lightpink,
        scriptscale=1.5, 
        width=2.857142857142857,
        height=1,
        depth=4,
        border=none
        }
    };

\draw [connection] (expand-right) -- (1.5,0,0) node{\midarrow};

\coordinate [shift={(10, 2, 0)}] (sumhead) at (l1relu-right);
\draw[connection] (l1relu-right) -- ++(1,0,0) -- ++(0,2,0) -- (sumhead) node{\midarrow} -- ++(0,-2,0);
\draw [connection] (expand-right) -- ++ (3.15,0,0) node{\elementsum} -- ++ (0,0,0);

\pic[shift={(1.5,0,0)}] at (-1.5,0,0) 
    {ConvReLU={
        name=l2+relu,
        opacity=\myopacity,
        caption=L$_4$,
        xlabel=$F$, 
        fill=lightpink,
        scriptscale=1.5, 
        width=1.4285714285714286,
        height=1,
        depth=4,
        border=none
        }
    };

\draw [connection] (l2+relu-right) -- ++ (1.5,0,0) node{\midarrow};

\pic[shift={(1.5,0,0)}] at (l2+relu-right) 
    {BaseBox={
        name=l3,
        opacity=\myopacity,
        caption=L$_5$,
        xlabel=$F$, 
        fill=salmon, 
        scriptscale=1.5, 
        width=1,
        height=1,
        depth=4,
        border=none
        }
    };

\coordinate [shift={(4, 2, 0)}] (sumhead) at (l2+relu-right);
\draw[connection] (l2+relu-right) -- ++(1,0,0) -- ++(0,2,0) -- (sumhead) node{\midarrow} -- ++(0,-2,0);
\draw [connection] (l3-right) -- ++ (3,0,0) node{\elementsum} -- ++ (0,0,0);

\draw [dotted, very thick] (6.9,0,0) -- ++ (1.6,0,0);
\draw [connection] (8.5,0,0) -- ++ (1.5,0,0);

\pic[shift={(6,0,0)}] at (l3-right) 
{ConvReLU={
		name=l10+relu,
		opacity=\myopacity,
		caption=L$_{12}$,
		xlabel=$F$, 
		fill=lightpink,
		scriptscale=1.5, 
		width=1.4285714285714286,
		height=1,
		depth=4,
		border=none
	}
};

\draw [connection] (l10+relu-right) -- ++ (1.5,0,0) node{\midarrow};

\pic[shift={(1.5,0,0)}] at (l10+relu-right) 
{BaseBox={
		name=l11,
		opacity=\myopacity,
		caption=L$_{13}$,
		xlabel=$F$, 
		fill=salmon, 
		scriptscale=1.5, 
		width=1,
		height=1,
		depth=4,
		border=none
	}
};

\coordinate [shift={(4, 2, 0)}] (sumhead) at (l10+relu-right);
\draw[connection] (l10+relu-right) -- ++(1,0,0) -- ++(0,2,0) -- (sumhead) node{\midarrow} -- ++(0,-2,0);
\draw [connection] (l11-right) -- ++ (3,0,0) node{\elementsum} -- ++ (0,0,0);

\pic[shift={(7,0,0)}] at (10,0,0) 
    {BaseBox={
        name=l12,
        opacity=\myopacity,
        caption=L$_{14}$,
        xlabel=$F$, 
        fill=salmon, 
        scriptscale=1.5, 
        width=1,
        height=1,
        depth=4,
        border=none
        }
    };

\draw[connection] (-.5, 0, 0) -- ++(0, 3, 0) -- node{\midarrow} ++(20,0,0) -- ++(0,-3,0);
\draw [connection] (l12-right) -- ++ (3,0,0) node{\elementsum} -- ++ (0,0,0);

\pic[shift={(3,0,0)}] at (l12-right) 
    {BaseBox={
        name=l13,
        opacity=\myopacity,
        caption=L$_{15}$,
        xlabel=$F$, 
        fill=salmon, 
        scriptscale=1.5, 
        width=1,
        height=1,
        depth=4,
        border=none
        }
    };

\draw [connection] (l13-right) -- ++ (1.5,0,0) node{\midarrow};

\pic[shift={(1.5,0,0)}] at (l13-right) 
    {SampleShuffle={
        name=ps,
        opacity=\myopacity,
        xlabel=$R$,
        fill=nephritis, 
        scriptscale=1.5, 
        height=1,
        width=.4,
        depth=4,
        border=none,
        dist=2,
        scalefactor=2,
        connectlineopacity=0.5,
        }
    };



\end{tikzpicture}}}
    \end{minipage}
    \caption{\textbf{Our proposed StofNet architecture} employs multiple convolutional layers and residual skip connections for improved feature representation. The model takes as input (light blue) a \mbox{1-D} signal with $C$ channels for optional feature concatenation. The initial layer (pink and orange) applies a \mbox{1-D} convolution with $F$ filters and a kernel size of 9 followed by a Rectified Linear Unit (ReLU). Layer 2 and 3 represent the semi-global bottleneck block consisting of \mbox{1-D} convolutions with a kernel size of 5, ReLUs, and a down- as well as upsampling block (purple and blue), respectively. The subsequent layers (4 to 14) consist of \mbox{1-D} convolutions with $F$ filters and a kernel size of 7. Residual connections are added after every other layer, whereas a ReLU (orange) follows convolutions without residual connections. The second last layer uses a \mbox{1-D} convolution with $F$ filters and a kernel size of 3, followed by an element-wise addition with the third layer residual output. The output is obtained by applying a \mbox{1-D} convolution with the specified upsampling factor $R$ and a kernel size of 3, followed by a sample shuffling operation (green).}
    \label{fig:arch}
\end{figure*}

\section{Method}
\label{sec:method}
%

%
For an input signal $\mathbf{X}\in\mathbb{R}^{N\times C}$ of $N$ samples and $C$ features, we wish our network \mbox{$f(\mathbf{X}):~\mathbb{R}^{N\times C}\mapsto\mathbb{R}^{NR}$} to learn how to discover local peaks using trained weights. 
We develop our deep neural network along the lines of \mbox{super-resolution} techniques in the computer vision field by using stacked residual blocks followed by upscaling through learned feature channel  shuffling~\cite{shi2016real,lim2017enhanced}. 
As we seek a localization method with refinement ability even in large contextual regions, we introduce a single bottleneck block for semi-global context recognition at an early stage of our proposed architecture. Contrary to~\cite{Zonzini:2022,kuleshov2017audio,ravanelli2018speaker}, our model omits a global bottleneck contraction to retain crucial resolution information in the temporal domain. A schematic visualization of our network is depicted in Fig.~\ref{fig:arch}. While our proposed model accepts varying input lengths with parameterizable block dimensions, we follow the image \mbox{super-resolution} field with $F=64$ feature channels and an upsampling factor of $R=4$.

\noindent\textbf{Loss.} Similar to previous work~\cite{liu2020deep}, we construct a binary mask \mbox{$\mathbf{y}\in\{0,1\}^{NR}$} where each 1 represents a ground truth sample position. We convolve this binary mask using a \mbox{1-D} Gaussian kernel $\mathbf{g}_{\sigma}\in\left[0,1\right)^{7}$ with $\sigma=1$
to learn the prediction of smooth local maxima distributions. The loss function reads,
\begin{align}
\mathcal{L}(\mathbf{X},\mathbf{y})= \lVert f(\mathbf{X}) - \lambda_0(\mathbf{g}_{\sigma} \circledast \mathbf{y})\rVert_2^2 + \lambda_1 \lVert f(\mathbf{X})\rVert_1
\end{align}
where $\circledast$ denotes the convolution operator, $\lambda_0$ amplifies the labels and $\lambda_1$ scales an $L_1$ regularization term that prevents $f(\cdot)$ from predicting an excessive amount of false positives. 

\noindent\textbf{Training.} Each model is trained using an AdamW optimizer with batch size 4, weight decay of $1\mathrm{e}{-8}$ and a start learning rate of $5\mathrm{e}{-4}$ using cosine annealing as a scheduler. The regularization scales are set to $\lambda_0=(\max(\mathbf{g}_{\sigma}\circledast\mathbf{y})/20)^{-1}$ and $\lambda_1=1\mathrm{e}{-2}$. We train a network for a maximum of 80 epochs with early stopping using a delta of $1\mathrm{e}{-6}$ and a patience of 5. \par
For better generalization, we prevent \mbox{over-fitting} through augmentation of the input signals. This involves an amplitude normalization such that \mbox{$\mathbf{X}\in\left[-1,1\right]^{N\times C}$} and the addition of noise for the training frames with a signal-to-noise ratio of $30~\text{dB}$. For learning single echo detection, we apply a randomized positional frame cropping to $\sfrac{3}{4}\,N$ samples in the temporal domain with subsequent padding for consistent frame length $N$ as required by baseline models~\cite{kuleshov2017audio,ravanelli2018speaker}.

\noindent\textbf{Inference.} After training, a single target localization is accomplished via ${\operatorname{arg\,max}} \{f(\mathbf{X})\}$. For multiple localizations, we employ thresholding in conjunction with non-maximum suppression. The ideal threshold depends on the network and dataset and is determined by Receiver Operating Characteristic (ROC) curve analysis using the \mbox{geometric-mean}~\cite{kubat1997addressing}.
\section{Experiments}
\label{sec:results}
For the evaluation of state-of-the-art ToF detection methods, we conduct a benchmark comparison on two different datasets with suitable metrics that are explained hereafter. The code of the following analyses, including backbone models, can be found online\footnote{Access to our code repository at \href{https://github.com/hahnec/stofnet}{https://github.com/hahnec/stofnet}\label{foot:code}}.

\begin{figure}[h!]
    \centering
    \includegraphics[width=.5\linewidth]{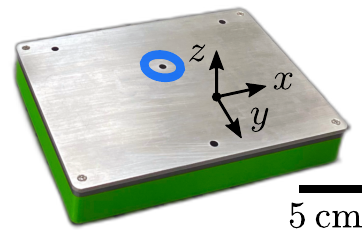}
    \caption{\textbf{Our prototype} uses the
\mbox{DK-CH101} airborne ultrasound sensor (blue circle) to acquire the \textit{SToF-Chirp} dataset\footref{foot:chirp}.}
    \label{fig:proto}
\end{figure}

\noindent\textbf{Datasets.} For numerical evaluation, we employ the PALA dataset\footnote{Access to PALA dataset at \href{https://doi.org/10.5281/zenodo.4343435}{doi.org/10.5281/zenodo.4343435}\label{foot:pala}} published by Heiles \textit{et al.}~\cite{heiles2022pala} as it provides in silico A-scans coming with ground truth data from third-party software that meets industrial standards. We customize settings for PALA by interpolating In-phase and Quadrature (IQ) signals by factor 20 and use $\lambda_1=1$ for training. 
%
Besides, we employ and release the SToF-Chirp dataset\footnote{Access to our SToF-Chirp dataset at \href{https://doi.org/10.5281/zenodo.8255670}{doi.org/10.5281/zenodo.8255670}\label{foot:chirp}}. The dataset was collected using our prototype depicted in Fig.~\ref{fig:proto}, which carries a TDK \mbox{DK-CH101} Smartsonic airborne ultrasound transducer operating at 175~kHz. We demodulate the IQ samples with an interpolation factor of 10.  
The dataset comprises three target classes, namely: wooden ruler, metal stick, and paper chart. The targets were moved along the $z$ direction with a total of nine distance measurements captured for each class. 
We collected 100 frames per class and distance, with 20 measurements designated for testing purposes and 80 measurements for training. 
The determination of the true distance for each measurement is based on the average of the manufacturer's estimates from 200 separate frames (distinct from test and train sets), providing a reliable reference for evaluating localization accuracy.
%

\noindent\textbf{Metrics.} For localization assessment, the minimum Root Mean Squared Error (RMSE) is calculated from the estimated position and the nearest ground truth position. Here, only RMSEs less than a tolerance $\tau$ are considered true positives and contribute to the total RMSE of all frames. In cases where the RMSE distance exceeds the tolerance, an estimated position is considered a false positive. As a result, ground truth locations without an estimate within the $\tau$ neighborhood are treated as false negatives. The Jaccard Index is employed to evaluate the ToA detection reliability, which takes into account both true positives and false negatives, offering a robust measure of each algorithm's performance. In addition, we provide the single batch inference time and parameter number for each model. 

\noindent\textbf{Baselines.} Our benchmark comparison comprises a gradient-based peak detection~\cite{Hahne:22}, the audio super-resolution network by Kuleshov~\textit{et al.}~\cite{kuleshov2017audio}, the ToA-tailored architectures by Zonzini~\textit{et al.}~\cite{Zonzini:2022}, the SincNet~\cite{ravanelli2018speaker} and \mbox{1-D} equivalents of the much celebrated ESPCN~\cite{shi2016real} and EDSR~\cite{lim2017enhanced} nets. 
Zonzini~\textit{et al.}~\cite{Zonzini:2022} propose a large and a small architecture for embedded devices. While we employ the large model for the PALA dataset, the number of its contracting downsample layers cannot handle the frame lengths of the SToF-Chirp dataset, for which we utilize the small Zonzini model in this study. All models are trained and executed on an Nvidia GeForce RTX 3090 Ti.

\begin{figure*}[h]
    \centering
    \includegraphics[width=\linewidth]{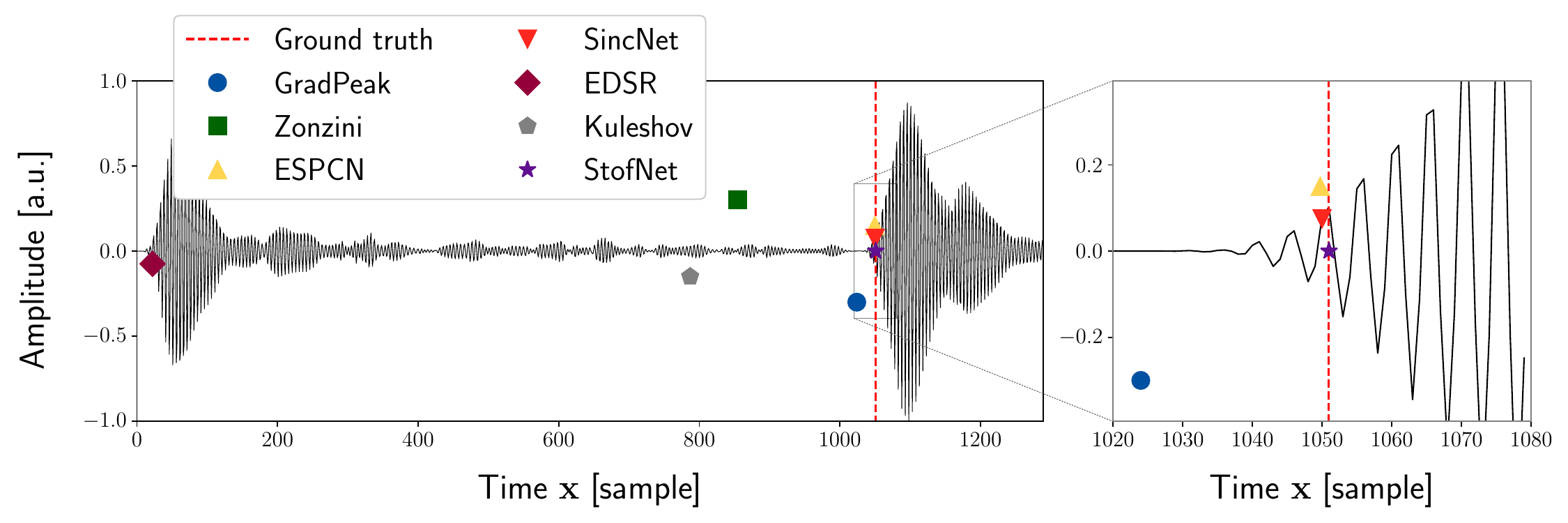}
    \caption{\textbf{Single echo detection plot} from the SToF-Chirp test set with a magnified view of the onset target position on the right.}
    \label{fig:chirp}
\end{figure*}
\begin{figure*}[h]
    \centering
    \includegraphics[width=\linewidth]{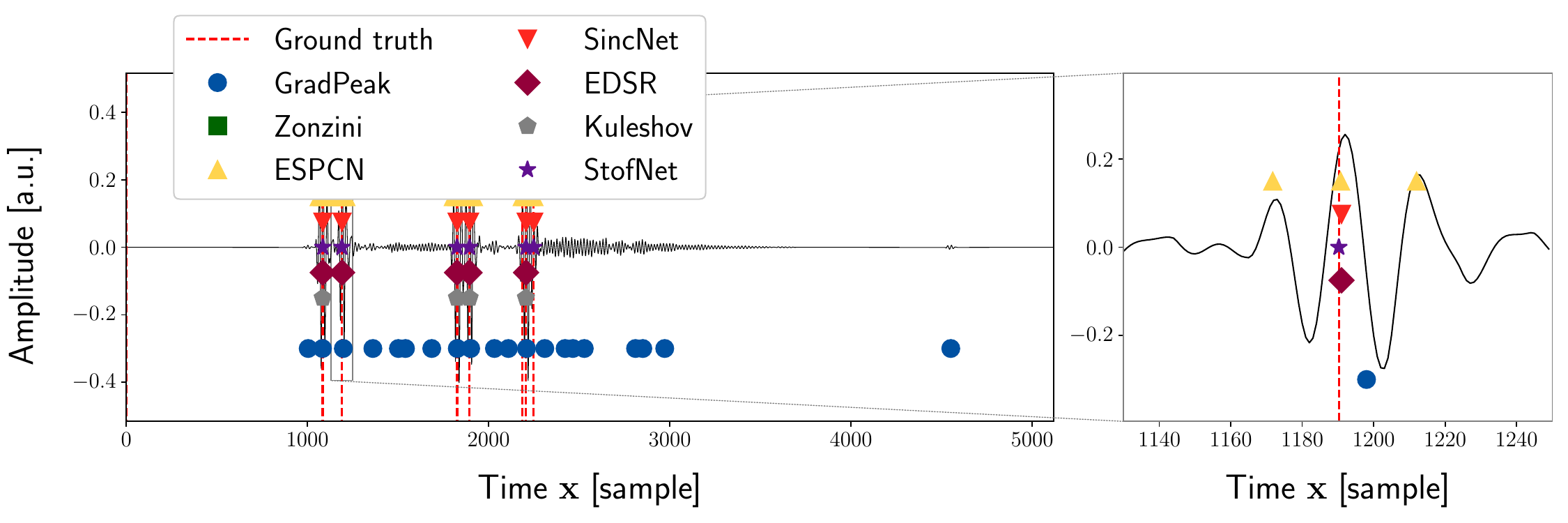}
    \caption{\textbf{Multiple echo detection plot} from the PALA test set with a magnified view of the peak target position on the right.}
    \label{fig:pala}
\end{figure*}

\noindent\textbf{Results.} For visual inspection, Figs.~\ref{fig:chirp} and \ref{fig:pala} depict exemplary channel data frames with localization results. %
The magnified portion in Fig.~\ref{fig:pala} reveals that ESPCN~\cite{shi2016real}, as a standalone super-resolution network, tends to detect multiple ToAs for a single target reflection. This observation serves as the motivation for introducing the Semi-Global (SG) context block in our StofNet architecture, facilitating our network to effectively investigate broader surroundings from condensed temporal information. 
To assess the effectiveness of the SG module, we conduct an ablation analysis, which corresponds to the absence of $\text{L}_2$ and $\text{L}_3$ in Fig.~\ref{fig:arch}.

Tables~\ref{tab:chirp} and \ref{tab:pala} list numerical results of the selected metrics for the SToF-Chirp and PALA dataset, respectively. \par
%
\begin{table}[h]
    \centering
    {
    \small
    \caption{Single echo detection results from SToF-Chirp data\footref{foot:chirp}. Units are provided in brackets. The RMSE tolerance is $\tau=1$.}
    \label{tab:chirp}
    \begin{tabularx}{\linewidth}{ 
         >{\raggedright\arraybackslash}p{5.2em} 
         >{\centering\arraybackslash}p{5.76em} 
         S[table-format=2.3] 
         >{\raggedleft\arraybackslash}p{2.8em} 
         >{\centering\arraybackslash}p{2em} 
         }
\toprule
\centering Model & {RMSE [Sample]} & {Jaccard~[\%]}& \centering Weights [k\#]  & Time [ms] \\
\midrule
Gradient~\cite{Hahne:22} & 0.607 $\pm$ \mbox{~~~n.a.} & 0.185 & 0 & 3.590 \\
Zonzini~\cite{Zonzini:2022} & 0.495 $\pm$ 0.161 & 0.370 & 134 & 2.159 \\
ESPCN~\cite{shi2016real} & 0.494 $\pm$ 0.295 & 21.111 & 6 & 1.671 \\
SincNet~\cite{ravanelli2018speaker} & 0.359 $\pm$ 0.282 & 70.741 & 329 & 2.879 \\
EDSR~\cite{lim2017enhanced} & 0.329 $\pm$ 0.297 & 51.481 & 210 & 2.811 \\
Kuleshov~\cite{kuleshov2017audio} & 0.229 $\pm$ 0.139 & 32.037 & 57416 & 3.267 \\
\hline
Ours w/o SG & 0.180 $\pm$ 0.217 & 65.370 & 318 & 2.540 \\
Ours w/ SG & 0.145 $\pm$ 0.212 & 83.519 & 645 & 3.273 \\
\bottomrule
\end{tabularx}
    }
\end{table}
%
In relation to RMSE, our StofNet model achieves the overall most precise localization in comparison to the other models. As for the Jaccard index, our model attains the highest score, indicating strong alignment in accurately detecting valid echoes within the defined accuracy threshold~$\tau$. 
All models exhibited relatively similar computational performance with inference time in the milliseconds range. 

%
\begin{table}[h]
    \centering
    {
    \small
    \caption{Multiple echo detection results from PALA data\footref{foot:pala}. Units are provided in brackets. The RMSE tolerance is $\tau=1$.}
    \label{tab:pala}
    \begin{tabularx}{\linewidth}{ 
         >{\raggedright\arraybackslash}p{5.2em} 
         >{\centering\arraybackslash}p{5.76em} 
         S[table-format=2.3] 
         >{\raggedleft\arraybackslash}p{2.8em} 
         >{\centering\arraybackslash}p{2em} 
         }
\toprule
\centering Model & {RMSE [Sample]} & {Jaccard~[\%]}& \centering Weights [k\#]  & Time [ms] \\
\midrule
Gradient~\cite{Hahne:22} & 0.532 $\pm$ 0.242 & 2.637 & 0 & 12.46 \\
Zonzini~\cite{Zonzini:2022} & 0.505 $\pm$ 0.290 & 0.151 & 1259 & 2.236 \\
ESPCN~\cite{shi2016real} & 0.405 $\pm$ 0.082 & 38.161 & 6 & 0.914 \\
SincNet~\cite{ravanelli2018speaker} & 0.389 $\pm$ 0.074 & 75.678 & 329 & 3.814 \\
EDSR~\cite{lim2017enhanced} & 0.287 $\pm$ 0.090 & 72.176 & 210 & 3.695 \\
Kuleshov~\cite{kuleshov2017audio} & 0.462 $\pm$ 0.145 & 34.858 & 516631 & 4.345 \\
\hline
Ours w/o SG & 0.266 $\pm$ 0.089 & 76.202 & 318 & 1.131 \\
Ours w/ SG & 0.253 $\pm$ 0.085 & 79.811 & 645 & 3.430 \\
\bottomrule
\end{tabularx}
    }
\end{table}
%
%
The super-resolution network proposed by Kuleshov~\textit{et al.}~\cite{kuleshov2017audio} is a heavyweight with respect to its 517M trainable weights. This network may serve well for spectral extrapolation as required by processing of audible signals, however, the imbalanced ratio of RMSE and Jaccard versus weight number in Table~\ref{tab:pala} makes it an unsuitable choice for ToF detection. 
One may note the difference in model weights for Kuleshov~\cite{kuleshov2017audio} and Zonzini~\cite{Zonzini:2022} in Tables~\ref{tab:chirp} and \ref{tab:pala}, which are due to varying input lengths of the different datasets.

The comparison demonstrates that our StofNet model offers an ideal balance in terms of precision (RMSE), reliability (Jaccard) and complexity (time and weights) underpinning its benefits and potential for practical metrology applications.
\section{Conclusion}
%
This study explored the potentials and limitations of modern super-resolution techniques for accurate, fast and reliable Time of Flight (ToF) sensing through a benchmark comparison from publicly available datasets. Overall, our proposed \mbox{1-D} localization network exhibits best performance compared to state-of-the-art attempts owing to its ability to extract high temporal fidelity while simultaneously building sufficient global context for more reliable predictions.
%
%
Moving forward, future research may focus on expanding the application of super-resolution networks to address specific challenges in deep learning such as domain gap across sensors and environmental conditions.
%
In summary, our study has shed light on the potential of modern super-resolution networks and their significant role in improving ToF sensing. The findings presented here provide valuable insights for researchers and practitioners in the field, paving the way for enhanced depth sensing performance in various domains.
%
\bibliographystyle{IEEEbib}
\bibliography{main}

\end{document}